\documentclass{article}

\PassOptionsToPackage{round}{natbib}
\usepackage[preprint]{neurips_2019}

\usepackage[utf8]{inputenc} % allow utf-8 input
\usepackage[T1]{fontenc}    % use 8-bit T1 fonts
\usepackage{hyperref}       % hyperlinks
\usepackage{url}            % simple URL typesetting
\usepackage{booktabs}       % professional-quality tables
\usepackage{amsfonts}       % blackboard math symbols
\usepackage{nicefrac}       % compact symbols for 1/2, etc.
\usepackage{microtype}      % microtypography
\usepackage{graphicx}
\usepackage{amsmath}
\usepackage{enumitem}
\usepackage{wrapfig}
\usepackage[bottom]{footmisc}

\usepackage{amsmath,amsthm,amssymb}
\usepackage{times}
\usepackage[font=footnotesize,labelfont=footnotesize]{subcaption}
\usepackage[font=footnotesize,labelfont=footnotesize]{caption}

\title{Learning Transferable Cooperative Behavior in Multi-Agent Teams}

\author{%
  Akshat Agarwal\thanks{Equal Contribution. Correspondence to \href{mailto:agarwalaks30@gmail.com}{agarwalaks30@gmail.com}, \href{mailto:sumit.sks4@gmail.com}{sumit.sks4@gmail.com} }\\
  Carnegie Mellon University\\
  Pittsburgh, USA \\
  \And
  Sumit Kumar\footnotemark[1]\\
  Carnegie Mellon University\\
  Pittsburgh, USA\\
  \AND
  Katia Sycara \\
  Carnegie Mellon University\\
  Pittsburgh, USA\\
}

\begin{document}

\maketitle

\begin{abstract}
While multi-agent interactions can be naturally modeled as a graph, the environment has traditionally been considered as a black box. We propose to create a shared agent-entity graph, where agents and environmental entities form vertices, and edges exist between the vertices which can communicate with each other. Agents learn to cooperate by exchanging messages along the edges of this graph. Our proposed multi-agent reinforcement learning framework is invariant to the number of agents or entities present in the system as well as permutation invariance, both of which are desirable properties for any multi-agent system representation. We present state-of-the-art results on coverage, formation and line control tasks for multi-agent teams in a fully decentralized framework and further show that the learned policies quickly transfer to scenarios with different team sizes along with strong zero-shot generalization performance. This is an important step towards developing multi-agent teams which can be realistically deployed in the real world without assuming complete prior knowledge or instantaneous communication at unbounded distances.
\end{abstract}

\section{Introduction}

Cooperative multi-agent systems find applications in domains as varied as telecommunications, resource management and robotics, yet the complexity of such systems makes the design of heuristic behavior strategies difficult. 
While multi-agent reinforcement learning (MARL) enables agents to learn cooperative behavior to maximize some team reward function, it poses significant challenges including the non-stationarity of the environment, combinatorially growing joint action and state spaces of the agents, and the multi-agent credit assignment problem.
Practically, most real world environments have partial observability (due to limited range and/or noisy sensors) and limited communication, which means agents have to learn to behave cooperatively conditioned only on local observations and limited communication.

While multi-agent systems have been modeled as graphs in previous works \citep{sukhbaatar2016learning,hoshen2017vain}, the environment has been usually treated as a black box. The agents receive information about other agents and entities in the environment in the form of a single vector or image with everything stacked together, which is a gross under-utilization of the natural structure present in the environment. Here, we propose to incorporate the inherent high-level structure of the environment directly in the learning framework by creating a shared agent-entity graph where both, agents and environmental entities, form vertices and edges exist between those vertices whose occupants can communicate with each other. Agents learn to achieve global consensus important for solving fully cooperative tasks by sending and receiving messages along the edges of this graph \citep{scarselli2009graph,gilmer2017neural}. 

Building on the framework of Graph Neural Networks \citep{vaswani2017attention, jiang2018graph}, we propose a multi-agent reinforcement learning (MARL) model that is invariant to the number of agents or entities present in the environment, and also invariant to the order or permutation of entities. This facilitates transferring policies trained for one team in a specific environment to a team with different number of agents and/or an environment with a different number of entities. We further show that the team of agents can learn complex cooperative strategies via a curriculum of progressively increasing difficulty. To the best of our knowledge, this is the first work which addresses the problem of multi-agent transfer and curriculum learning of cooperative behaviors in a decentralized framework. 

\begin{figure*}[ht]
    \centering
    \includegraphics[width=.8\textwidth]{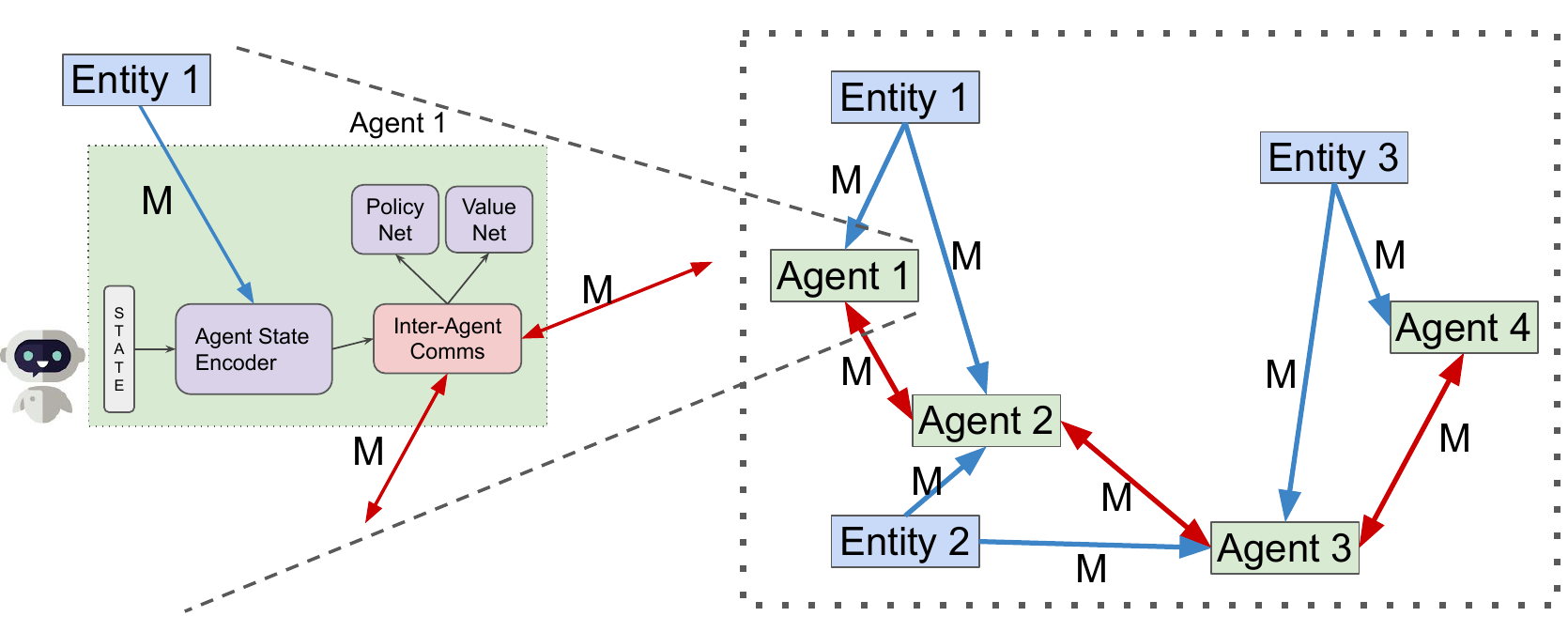}
    \caption{The proposed shared agent-entity graph on the right, and a detailed look at the internal architecture of each agent on the left. Messages exchanged between agents are depicted by red edges while those between an entity and an agent are shown by blue edges.} 
    \label{fig:overall}
\end{figure*}

\section{Related Work}
MARL has been a widely studied topic in the machine learning community. One of the earliest works, Independent Q-learning \citep{tan1993multi,tampuu2017multiagent} trains independent Q-value functions for each agent using regular Q-learning \citep{watkins1992q}, while assuming that the other agents are a part of the environment. Since the other agents are also learning, the environment becomes non-stationary and the resulting instability prevents these methods from scaling to more than $2$ agents. 

Under the paradigm of centralised learning with decentralised execution, a multitude of recent works have trained actor-critic algorithms where the critic is centralised and makes use of global information available during training. During execution, however, agents only use their actor network for selecting actions which enables the entire system to operate in a fully decentralized manner. 
MADDPG \citep{lowe2017multi} learns a centralised critic for each agent by providing the joint state and actions of all agents to the critic, and trains policies for each agent using the DDPG \citep{lillicrap2015continuous} algorithm. 
COMA \citep{foerster2018counterfactual} also uses a centralised critic but estimates a counterfactual advantage function that helps with multi-agent credit assignment by isolating the effect of each agent's action. VDN \citep{sunehag2017value} decomposes a centralized state-action value function into a sum of individual agent specific functions. The decomposition imposes a strict prior which is not well justified, and limits the complexity of the agents' learned value functions. 
Q-Mix \citep{rashid2018qmix} improves upon this by removing the requirement of additive decomposition of the centralised critic, instead imposing a less restrictive monotonicity requirement on agents' individual state-action value functions, and allowing a learnable mixing of the individual functions which does not limit the complexity of functions that could be learned. 
All of these works, however, do not utilize any structure present in the environment. Instead each agent's observation is simply a concatenation of the states of other agents and various features of the environment. 
Also, the use of a centralized critic prevents the learned policy to generalize to scenarios with different number of agents in the team than the one during training. Moreover, lack of communication between agents prevents these methods to be successful in scenarios where explicit coordination is required to solve the task, as we demonstrate in this work. 

CommNet \citep{sukhbaatar2016learning} is one of the earliest works to learn a differentiable communication protocol between multiple agents in a fully cooperative centralized setting. However, they did not explicitly model interactions between agents, instead each agent receives the averaged states of all its neighbors. VAIN \citep{hoshen2017vain} improves upon the mean aggregation by using an exponential kernel based attention to selectively attend to the messages received from other agents, and showed predictive modeling of multi-agent systems using supervised learning. In this work, we use the scaled dot-product attention mechanism proposed by \cite{vaswani2017attention} for inter-agent communication, which can be easily substituted with the ones used in CommNet and VAIN.

% discrete communication channel
\cite{foerster2016learning} demonstrated end-to-end learning of protocols in complex environments inspired by communication riddles and multi-agent computer vision problems with partial observability. \cite{mordatch2018emergence} also demonstrated emergence of compositional language in multi-agent systems in both cooperative and competitive settings. They, however, learned discrete communication using symbols from a limited vocabulary, and made it end-to-end differentiable by using the Gumbel-softmax estimator. In contrast, our communication module is continuous and fully differentiable. 

In work done concurrently to ours, TarMAC \citep{das2018tarmac} uses dot-product attention mechanism for inter-agent communication, however, they do not impose any restrictions on communication, leading to a centralized execution paradigm. DGN \citep{jiang2018graph} also uses a similar mechanism for communication but with Q-learning \citep{mnih2013playing} for training. They allowed each agent to communicate with its $3$ closest neighbors. From a practical consideration, communication between agents is usually restricted by their mutual separation, meaning that an agent can communicate only with neighbors within a certain radius. We would also like to emphasize that being able to communicate with the $3$ closest neighbors ensures that the agents' graph is always a single connected component and no agents are ever disconnected from the others, while having a distance-based restriction leads to formation of several different connected components in the agents' graph, none of which can communicate with each other leading to a significantly more difficult learning (to cooperate) problem. 

\section{Method}

\subsection{Agent-Entity Graph}
\label{sec:graph}
An environment can often be described as a set of different entities with a defined structure. For example, the environment for a self-driving vehicle includes other vehicles, traffic lights, pedestrians, etc. which are interacting with each other. Also, for multi-agent systems or swarms, the environment can be represented as a set of obstacles and/or landmarks. Instead of treating the environment as a black-box, we propose to utilize the inherent high-level structure in the learning process itself.

We define a graph $\mathcal{G} := (\mathcal{V},\mathcal{E})$ where each node $n \in \mathcal{V}$ is either an agent or an environment entity, and there exists an edge $e \in \mathcal{E}$ between two nodes if the node occupants can communicate with each other. In this work, we consider static entities, i.e., their positions remain same throughout an episode. However, across different episodes, the entities can take random positions in the environment. Also, we assume that the agents have access to the position of all the entities at the beginning of each episode. This means that there always exists an edge between each agent-entity pair. With respect to communication between agents, we consider two variants: 

\noindent \textbf{Restricted Communication:} Two agents can communicate with each other only if they are separated by a distance less than a pre-defined threshold (or communication bandwidth). 

\noindent \textbf{Unrestricted Communication:} All agents can communicate with each other. In this case, $\mathcal{G}$ is a fully-connected graph. 

Modeling the multi-agent system as a graph provides a strong inductive bias to the learning algorithm. In practice, agents which are closer to each other have a greater impact on each others' behavior, and this crucial information is baked into the graph architecture itself, which greatly aids learning. 

\subsection{Learning to communicate} \label{sec:comms}
We now describe the message passing mechanism by which agents establish a global consensus among themselves in order to accomplish the given task. Each agent $i \in \mathcal{V}$ observes only its own local state $X^i$. In this work, the state of an agent comprises of its position and velocity. The agent forms its state encoding $U^i = f_a(X^i)$ by using a learnable differentiable encoder network $f_a$. 

The agent then aggregates all the information about the environment into a fixed size embedding $E^i$ by using a Graph Neural Network (GNN). Specifically, it first forms an embedding $e^l_i = f_e(X^l_i)$ for each of the entities $l \in \mathcal{V}$ using an entity encoder function $f_e$. Here, $X^l_i$ is the state of entity $l$ w.r.t. agent $i$. Since entities are static, $X^l_i$ is simply the position of entity $l$ w.r.t. agent $i$. 
The agent then uses the dot product attention mechanism proposed by \cite{vaswani2017attention} to update the entities' embeddings $e^l_i$ and finally aggregate them together into a fixed size environment embedding $E^i$. We refer to this process as \textit{entity message passing}. Note that, there is no actual message transmission between entities and agents, but, the agents themselves do all the computation with the knowledge of entities' states. 

% why fixed size is important ? 
This environment information aggregation is an important step as it enables the agents to form a fixed size representation of the environment irrespective of the number of entities. Having a fixed size representation is important for being able to handle scenarios with different number of entities and also for transferring policies learned in one scenario to another. 
Other ways of forming a fixed size representation is to multi-channel grids or feature maps~\citep{resnick2018pommerman}, however, that unnecessarily increases the observation space of the agent and hence the complexity of the problem. 
Another common method is to take a conservative approach and allocate a vector of size more than required and pad the empty slots with some constant value like $0$ or $-1$ \citep{foerster2018counterfactual}. However, that requires a prior knowledge of an upper bound on the number of entities. Moreover, stacking all the observations in a single vector discards the high-level structural information present in the environment resulting in poor performance, as we show later in this work. 

It is often the case that a particular representation of the environment is suitable for a given task. For instance, consider a simple task where a team of $3$ agents is required to cover a \textit{set} of $3$ locations in such a manner that each agent goes to a distinct location. Clearly, in this case, the environment representation should be invariant to the permutation order of entities which is not be possible if one simply stacks all the entities' positions. Instead, the inherent structure of the environment can be incorporated in the learning framework itself by using a graph architecture as done in this work.

\begin{wrapfigure}{r}{0.45\textwidth}
  \begin{center}
    \includegraphics[width=0.45\textwidth]{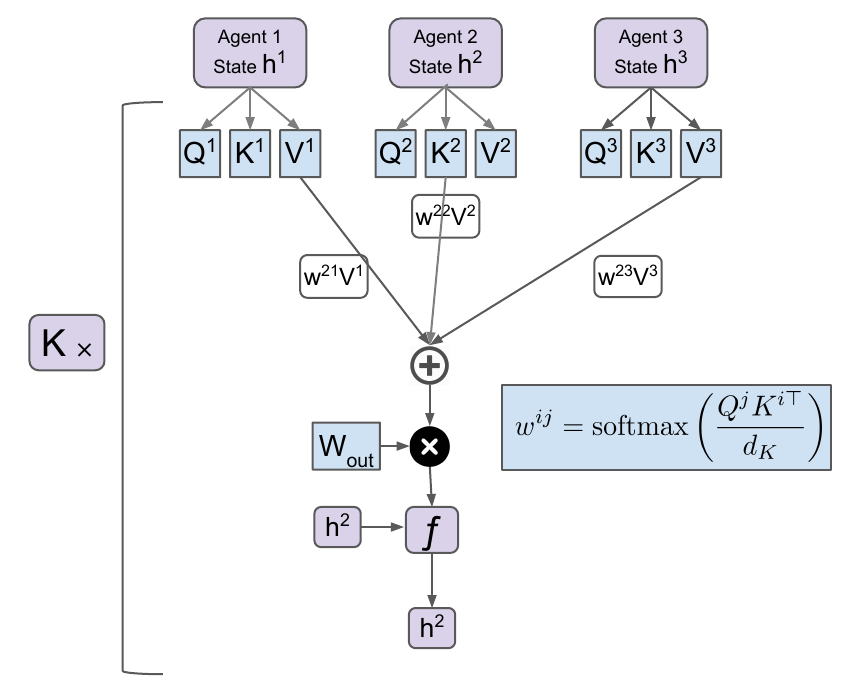}
  \end{center}
  \caption{Scaled dot-product attention mechanism for message passing.}
  \label{fig:message_passing}
\end{wrapfigure}

\noindent \textbf{Inter-agent communication: } After computing its state encoding $U^i$ and environment encoding $E^i$, agent $i$ concatenates them together into a joint encoding $h^i$. This encoding represents the agent's understanding of its own state and the environment. So far, the agent does not possess any information about its teammates. Now, each agent $j \in \mathcal{V}$ computes a key $K^j = W_K h^j$, query $Q^j=W_Q h^j$ and value $V^j = W_V h^j$ vectors where $W_K, W_Q$ and $W_V$ are learnable parameters. Agent $i$, after receiving query-value pair $(Q^j, V^j)$ from all of its neighbors $j \in \mathcal{N}(i)$, assigns weight $w^{ij} = \texttt{softmax} \left( \frac{Q^j K^{i \top} }{d_K}\right)$ to each of the incoming messages. Here, $d_K$ is the dimensionality of key vector. It then aggregates all the messages by computing a weighted sum of its neighbors' values followed by a linear transformation $V_f^i = W_{\text{out}} \sum w^{ij} V^j$ where $W_{\text{out}}$ is another learnable parameter. Finally, the agent updates its encoding by doing a non-linear transformation of its current embedding $h^i$ concatenated with $V_f^i$ by using a neural network $f$. We summarize our inter-agent communication module in Fig. \ref{fig:message_passing}.

The attention mechanism, described above, enables the agents to selectively attend to messages coming from its neighbors. Since the agent network may be sparsely connected with long chains in each connected component, we use multi-hop communication ($K$ rounds of message passing) to allow information to propagate between agents that might not be directly connected with each other. 

After $K$ rounds of message passing, each agent has an updated encoding $h^i$. It then feeds this encoding into another neural network with value and policy heads to predict an estimate of its state value and a probability distribution over all possible actions respectively. Each agent samples an action from the distribution and acts accordingly, upon which the environment gives a joint reward to the team. In this work, we consider scenarios where the agents form a homogeneous team and share all the learnable parameters including those of agent encoder network, entity encoder network, graph networks, and policy and value networks. Since each agent receives different observations, attends incoming messages from other agents differently and perceives the environment differently (relative state of all the entities), sharing parameters does not preclude them from behaving differently, as is appropriate. The entire model is trained in an end-to-end manner using the actor-critic policy gradient PPO \citep{schulman2017proximal} algorithm. A salient feature of our proposed model is that it can be trained and executed both in a completely decentralized manner.

\subsection{Curriculum Training}
\label{sec:curriculum}
Since our model is invariant to the number of agents or entities, sharing network parameters among all the agents enables us to directly use a policy $\pi$ trained for a task $\mathcal{T}$ with $M$ agents and $L$ entities to a different task $\mathcal{T'}$ with $M'$ agents and $L'$ entities. The policy $\pi$ can serve as a good initialization for task $\mathcal{T'}$ which can be improved further by fine-tuning with some experiences collected in $\mathcal{T'}$. This facilitates in establishing a curriculum \citep{bengio2009curriculum} of tasks with increasing difficulty. Agents first learn cooperative behaviors in a small team and with the addition of new members bootstraps their strategies to accomplish the goal for this larger team. In other words, they utilize their previous knowledge in a new scenario and gradually learn complex cooperative strategies in a large team.

Curriculum learning with our proposed shared agent-entity graph enables us to train policies for complex tasks directly training on whom yields poor performance. Using a graph architecture is known to induce a strong inductive bias to the learning algorithm. To analyze this effect, we also evaluate the zero-shot generalization performance of our model. 

\section{Experiments}

\subsection{Task Description}
\begin{figure}[h]
    \centering
    \begin{minipage}{0.33\textwidth}
        \centering
        \frame{\includegraphics[width=\textwidth]{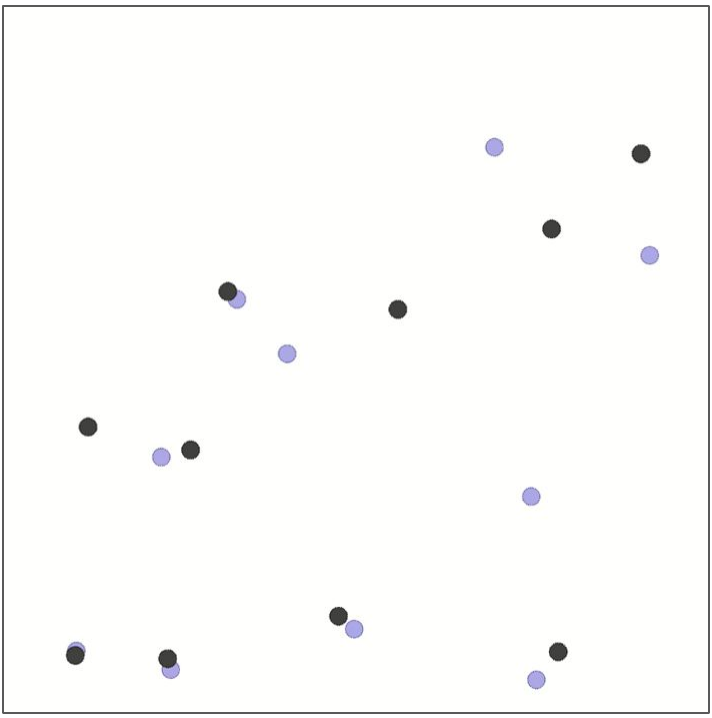}}
        \subcaption{Coverage Control}
        \label{fig:cc}
    \end{minipage}\hfill
    \begin{minipage}{0.33\textwidth}
        \centering
        \frame{\includegraphics[width=\textwidth]{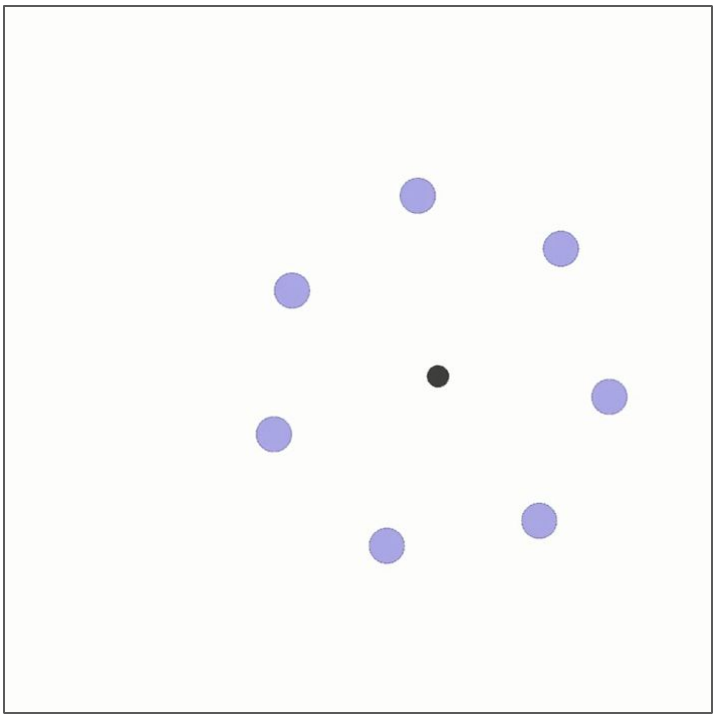}}
        \subcaption{Formation Control}
        \label{fig:fc}
    \end{minipage}
    \begin{minipage}{0.33\textwidth}
        \centering
        \frame{\includegraphics[width=\textwidth]{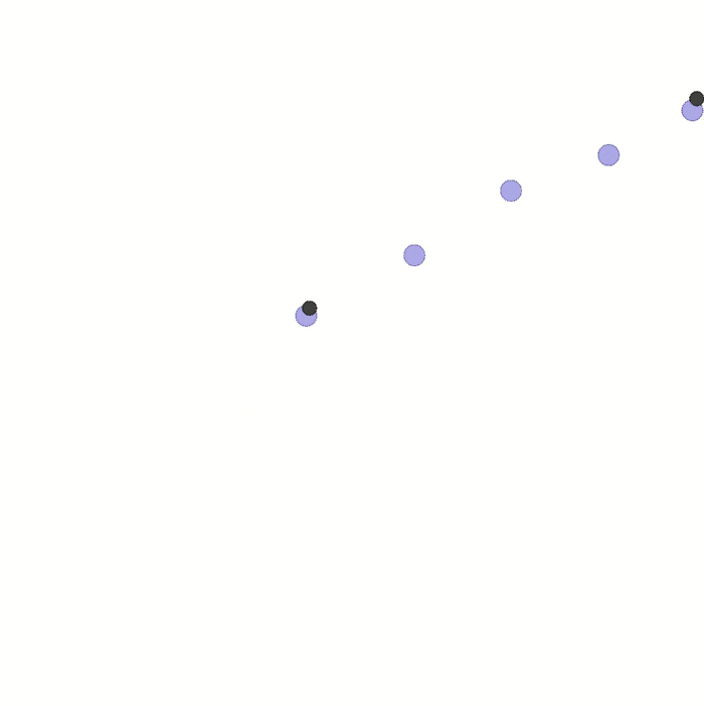}}
        \subcaption{Line Control}
        \label{fig:lc}
    \end{minipage}
    \caption{Simulation Environments used in this work. Agents are shown in blue circles while the landmarks in grey ones.}
\end{figure}

We evaluate our proposed model on three standard swarm robotics tasks \citep{mesbahi2010graph,balch1998behavior}: \textit{coverage control}, \textit{formation control} and \textit{line control}. We have implemented them in the Multi-Agent Particle Environment \footnote{\url{https://github.com/openai/multiagent-particle-envs}} \citep{lowe2017multi} where the agents can move around in a $2$D space following a double integrator dynamics model \citep{rao2001naive}. 
The action space for each agent is discretized, with the agent being able to control unit acceleration or deceleration in both X and Y directions. We briefly describe the three environments below:

\noindent \textbf{Coverage Control: } There are $M$ agents and $M$ landmarks in the environment (see Figure \ref{fig:cc}). The objective is for the agents to deploy themselves in a manner such that every agent reaches a distinct landmark. Note that we do not assign particular landmark to each agent, but instead let the agents communicate with each other and develop a consensus as to who goes where.

\noindent \textbf{Formation Control: } There are $M$ agents and $1$ landmark in this environment (see Figure \ref{fig:fc}). The agents are required to position themselves into an $M$-sided regular polygonal formation, with the landmark at its centre. 

\noindent \textbf{Line Control: } There are $M$ agents and $2$ landmarks in the environment (see Figure \ref{fig:lc}). The objective is for the agents to position themselves equally spread out in a line between the two landmarks. 

\subsection{Implementation Specifications}
The agent encoder $f_a$ and the entity encoder $f_e$ takes as input the $4$-dim agent states and $2$-dim entity states respectively and outputs a $128$-dim embedding. Both the encoders are a single ReLU fully connected (FC) layer. The communication module uses attention with $128$-dim queries, keys and values. The aggregated message is concatenated with the agent's state and passed through a single ReLU FC layer $f$ containing $128$ neurons as the update function. We use $K=3$ communication hops between the agents. Both the policy and value heads are $2$ ReLU FC layers with $128$ neurons. We use orthogonal initialization \citep{saxe2013exact} for all the learnable parameters. We have open-sourced our repository containing all the simulation environments and codes \footnote{\url{https://github.com/sumitsk/matrl.git}}.

All the environments are $2 \times 2$ sq. units in size as is the standard in MAPE. In the restricted communication version, we set the communication distance to be $1$ unit. Each episode lasts for a total of $50$ timesteps. Evaluation is carried out after every $50$ updates on $100$ episodes in a newly seeded environment. During evaluation, each agent performs greedy decentralized action selection. Each PPO update is performed after accumulating experience for $128$ timesteps on $32$ parallel processes, or equivalently, every $4096$ total timesteps.   
 
\subsection{Results}
\label{sec:results}
We used $3$ metrics to compare different methods: \textbf{Success Rate} (\texttt{S}$\%$): In what percentage of episodes does the team achieve its objective? (Higher is better) \textbf{Time} (\texttt{T}): How many time steps does the team require to achieve its objective? (Lower is better) \textbf{Average Distance} (\texttt{DIST.}): What is the average distance of a landmark from its closest agent? This metric is used in coverage control task only. (Lower is better).

We could not find any prior work on multi-agent reinforcement learning in coverage, formation or line control tasks and hence do not have previously published results to compare with. We used publicly available implementations\footnote{\url{https://github.com/oxwhirl/pymarl}, \url{https://github.com/openai/maddpg}} to compare with Q-Mix \citep{rashid2018qmix}, VDN \citep{sunehag2017value}, IQL \citep{tampuu2017multiagent}, COMA \citep{foerster2018counterfactual} and MADDPG \citep{lowe2017multi}. These methods rely on access to the global state of the system (for example, a centralized view of the entire system) during training instead of inter-agent communication for emergence of cooperative behaviors. For these methods, the agents have full observability, i.e., they know the position and velocity of all the other agents at every time step. In contrast, agents are unaware of the state of other agents in our method. The corresponding results are shown in Table \ref{tab:prevwork}. 

\begin{table}
\caption{Comparisons with prior works with $M=3$ and $M=6$ agents. \texttt{UC}: Unrestricted Communication, \texttt{RC}: Restricted Communication, \texttt{T}: Average Episode Length, \texttt{S}$\%$: success rate, \texttt{DIST}: average agent-landmark distance.} 
\label{tab:prevwork}
\vskip 0.15in
\begin{center}
\begin{small}
\begin{sc}
\begin{tabular}{lccccccccc}
\toprule
& & Observ- & & \multicolumn{3}{c}{$M=3$} & \multicolumn{3}{c}{$M=6$}  \\
Task & Method & ability & Comm & Dist. & T & S $\%$ & Dist. & T & S $\%$\\
\midrule
Coverage & Q-Mix  & Full & N/A & 0.46 & 50 & 0 & 0.51 & 50 & 0\\
Coverage & VDN    & Full & N/A & 0.44 & 50 & 0 & 0.47 & 50 & 0\\
Coverage & IQL    & Full & N/A & 0.51 & 50 & 0 & 0.43 & 50 & 0\\
Coverage & COMA   & Full & N/A & 0.41 & 50 & 0 & 0.43 & 50 & 0\\
Coverage & MADDPG & Full & N/A & 0.065 & 17.89 & 95  & 0.52 & 50 & 0\\
Coverage & Ours & Partial & UC & 0.047 & 14.12 & 100 & 0.15 & 20.47 & 93\\
Coverage & Ours & Partial & RC & 0.049 & 14.22 & 98  & 0.17 & 48.32 & 5 \\ % 202,197
\midrule
Formation & MADDPG & Full & N/A & -- & 15.66 & 100 & -- & 50 & 0\\
Formation & Ours & Partial & UC & -- & 13.56 & 100 & -- & 14.22 & 100 \\
Formation & Ours & Partial & RC & -- & 12.97 & 100 & -- & 14.26 & 100 \\
\midrule
Line & MADDPG & Full & N/A & -- & 35.84 & 58 & -- & 50 & 0\\
Line & Ours & Partial & UC & -- & 15.14 & 98 & -- & 16.31 & 100 \\
Line & Ours & Partial & RC & -- & 15.24 & 97 & -- & 17.07 & 100 \\
\bottomrule
\end{tabular}
\end{sc}
\end{small}
\end{center}
\vskip -0.1in
\end{table}

Even with full observability, only MADDPG is able to solve the coverage control and formation control tasks for $M=3$ agents. It is only partially successful in the line control task. All the other $4$ baseline methods report no success in the coverage control task. For the other two tasks too, the success rate is $0\%$, hence we do not report them in the table. On the other hand, our proposed method is able to solve all the given tasks even with partial observability. 

We also evaluated the models on a more challenging version of these tasks, with $M=6$ agents. In this case, all the baseline methods achieve no success at all. In contrast, our method is able to solve the formation control and line control task in both the communication versions. In the coverage control task, although it was not have been able to achieve near-perfect success rate, it still performed better than all the baselines. 

\subsection{Curriculum Training}
We observed from the last set of results that learning cooperative behaviors in a team becomes more challenging with increase in number of agents. Instead of training policies directly from scratch, we deploy a curriculum over the number of agents. A policy is first trained with $M=3$ agents. Once the team achieves a desired success rate threshold, the learned policy is then transferred to a team with $M=5$ agents. In other words, a team of $5$ agents start with the policy trained for $3$ agents. The training then begins for this team and on achieving the set threshold, the process is repeated with $7$ and finally $10$ agents. We set the success rate threshold to be $85\%$. 

In this work, we have incorporated entities and agents together in a shared graph and formed a fixed dimensional environment representation using entity message passing (EMP) mechanism as described in Section \ref{sec:comms}. As mentioned before, another commonly used alternative is to stack all the entities' state in a single vector and pad the vector with some constant value to make it some fixed size. We allocated a size of $20$ units, i.e., a maximum of $10$ landmarks and filled the slots corresponding to non-existent entities with $0$s. We refer to this approach as the one without EMP.  

\begin{table}[h]
\caption{Curriculum Learning for coverage control task. \texttt{EMP}: Entity Message Passing, \texttt{N}: Number of updates.}
\label{tab:cl_cc}
% \vskip -0.15in
\begin{center}
\begin{small}
\begin{sc}
\begin{tabular}{ccccccccccc}
\toprule
Comm & EMP & \multicolumn{2}{c}{$M=3$} & \multicolumn{2}{c}{$M=5$} & \multicolumn{2}{c}{$M=7$} & \multicolumn{2}{c}{$M=10$}\\
& & S$\%$ & N & S$\%$ & N & S$\%$ & N & S$\%$ & N\\
\midrule
No & UC & 92 & 2450 & 96 & 3900 & 0 & -- & -- & -- \\
No & RC & 90 & 2900 & 0 & -- & -- & -- & -- & -- \\
Yes & UC & 96 & 1100 & 92 & 250 & 98 & 1000 & 86 & 200 \\
Yes & RC & 91 & 1100 & 96 & 3700 & 81 & 50 & 85 & 3250 \\
\bottomrule
\end{tabular}
\end{sc}
\end{small}
\end{center}
\vskip -0.1in
\end{table}

We compare the performance of curriculum learning on coverage control tasks with and without EMP in the two communication scenarios. The results of which are shown in Table \ref{tab:cl_cc}. Looking at the performance of the model without EMP for $M=3$ agents, it is clear that increasing the size of observation space (by padding) increases the complexity of the problem and requires more samples to solve the task. Also, in the restricted communication version, the team is unable to accomplish the goal on addition of two members. With no restriction on communication, the team of $5$ agents do learn the desired behavior, however, no success is observed on moving to $7$. 

In contrast, our proposed model shows fast transfer across increasingly difficult tasks and ultimately, even a team with $10$ agents have learned the required cooperative strategies. Note that, the desired level of cooperation, even for a smaller team with $6$ agents, did not emerge when training was done directly from scratch in restricted communication setting (see Table \ref{tab:prevwork}). 

We also evaluated curriculum learning on the other two tasks of formation control and line control. Their results are shown in Table \ref{tab:curr_2}. In these environments too, our model shows efficient transfer across tasks and is able to instill optimal cooperative behaviors even in large teams. This shows that incorporating the structure of environment in the learning framework induces a strong inductive bias in the multi-agent system that aids in learning transferable skills.  

\begin{table}[h]
\caption{Curriculum Learning on formation control and line control tasks. \texttt{EMP}: Entity Message Passing, \texttt{N}: Number of updates.}
\label{tab:curr_2}
\vskip 0.15in
\begin{center}
\begin{small}
\begin{sc}
\begin{tabular}{cccccccccc}
\toprule
Task & Comm & \multicolumn{2}{c}{$M=3$} & \multicolumn{2}{c}{$M=5$} & \multicolumn{2}{c}{$M=7$} & \multicolumn{2}{c}{$M=10$}\\
& & S$\%$ & N & S$\%$ & N & S$\%$ & N & S$\%$ & N\\
\midrule
Formation & UC & 94 & 250 & 99 & 300 & 100 & 300 & 100 & 100\\
Formation & RC & 97 & 300 & 100 & 100 & 95 & 200 & 100 & 3100\\
\midrule
Line & UC & 86 & 1200 & 98 & 700 & 93 & 150 & 82 & 700\\
Line & RC & 93 & 1200 & 93 & 1450 & 88 & 3450 & 87 & 350\\
\bottomrule
\end{tabular}
\end{sc}
\end{small}
\end{center}
\vskip -0.1in
\end{table}

\subsection{Zero shot Generalization}
We evaluated the policy trained for $M=5$ agents directly without any fine-tuning on different team sizes. The corresponding results for all the three tasks are shown in Table \ref{tab:generalization}. In all the three tasks, the trained policy shows impressive zero-shot success rate in both the unrestricted and restricted communication settings. Such results show that our proposed model has been able to capture the inherent structure present in the environment, thanks to shared agent-entity graph architecture, and is able to solve tasks it has never seen before by utilizing its past experiences of related but different tasks.  

\begin{table}[h]
\caption{Zero Shot Generalization results. Policy trained for $M=5$ agents is evaluated directly for different team sizes without any fine-tuning and the obtained success rates (\texttt{S}$\%$) are reported.}
\label{tab:generalization}
\vskip 0.15in
\begin{center}
\begin{small}
\begin{sc}
\begin{tabular}{ccccccccccccc}
\toprule
Task & Comm & $M-3$ & $M-2$ & $M-1$ & $M=5$ & $M+1$ & $M+2$ & $M+3$\\
\midrule
Coverage & UC & 89 & 95 & 93 & 98 & 83 & 65 & 41\\
Coverage & RC & 84 & 92 & 99 & 99 & 99 & 95 & 74 \\
\midrule
Formation & UC & 1  & 9 & 98 & 100 & 91 & 21 & 1\\
Formation & RC & 1 & 68 & 99 & 99  & 34 & 30 & 8\\
\midrule
Line & UC & 0 & 0 & 57 & 99 & 81 & 45 & 16\\ 
Line & RC & 0 & 16 & 76 & 99 & 45 & 16 & 8\\
\bottomrule
\end{tabular}
\end{sc}
\end{small}
\end{center}
\vskip -0.1in
\end{table}

\section{Conclusion and Future Work}
Instead of treating the environment as a black box, we proposed to utilize the inherent structure in a shared agent-entity graph whose vertices are formed by both, the agents and environment entities. The agents learn cooperate behaviors by exchanging messages with each other along the edges of this graph. Our proposed model is invariant to the number of agents or entities present in the environment which enables us to establish a curriculum learning framework in multi-agent systems. We showed state-of-the-art results on coverage and formation control for swarms in a fully decentralized execution framework and demonstrated that the learned policies have strong zero-shot generalization to scenarios with different team sizes. We also showed that complex tasks, which are difficult to solve by directly training policies from scratch, can instead be solved via curriculum. 
For future work, we want to investigate the performance of the agent-entity graph when there is a team of evolving adversaries also in the environment. Also, developing curriculum learning algorithms for multi-agent teams in the presence of adversaries is another direction of research work from here.   
\bibliography{refs,SFrefs,references}
\bibliographystyle{plainnat}

\end{document}